\documentclass[letterpaper]{article}
\usepackage{aaai}
\usepackage{times}
\usepackage{helvet}
\usepackage{courier}
\usepackage{graphicx} 
\usepackage{subfigure}
\usepackage{epstopdf}
 
\begin{document}
%
\title{Exploring the Contribution of \\Unlabeled Data in Financial Sentiment Analysis}
\author{Jimmy SJ. Ren$^1$, Wei Wang$^1$, Jiawei Wang$^2$, Stephen Shaoyi Liao$^1$\\
$^1$Department of Information Systems, City University of Hong Kong, 83 Tat Chee Ave, Kowloon, Hong Kong\\
jimmy.sj.ren@gmail.com, \{wewang8, issliao\}@cityu.edu.hk\\
$^2$USTC-CityU Joint Advanced Research Centre, 166 Ren'ai Road, Suzhou, China\\
wangjw2@mail.ustc.edu.cn 
}

\maketitle
\begin{abstract}
\begin{quote}
With the proliferation of its applications in various industries, sentiment analysis by using publicly available web data has become an active research area in text classification during these years. It is argued by researchers that semi-supervised learning is an effective approach to this problem since it is capable to mitigate the manual labeling effort which is usually expensive and time-consuming. However, there was a long-term debate on the effectiveness of unlabeled data in text classification. This was partially caused by the fact that many assumptions in theoretic analysis often do not hold in practice. We argue that this problem may be further understood by adding an additional dimension in the experiment. This allows us to address this problem in the perspective of bias and variance in a broader view. We show that the well-known performance degradation issue caused by unlabeled data can be reproduced as a subset of the whole scenario. We argue that if the bias-variance trade-off is to be better balanced by a more effective feature selection method unlabeled data is very likely to boost the classification performance. We then propose a feature selection framework in which labeled and unlabeled training samples are both considered. We discuss its potential in achieving such a balance. Besides, the application in financial sentiment analysis is chosen because it not only exemplifies an important application, the data possesses better illustrative power as well. The implications of this study in text classification and financial sentiment analysis are both discussed.
\end{quote}
\end{abstract}

\section{Introduction}
Sentiment analysis (or opinion mining) is a research area with the goal of finding other people's opinions (Pang and Lee 2008). In recent years, many applications of sentiment analysis started to play an increasing important role in creating values for various industries. Such phenomenon should largely be attributed to the rapid development of the Web and the inevitable trend of Big Data. For instance, the number of product reviews in any electronic commerce website is ever increasing. Management and utilization of such reviews by sentiment analysis is becoming one of the key issues (Pang and Lee 2008). In financial applications, massive textual data in social networks enables the prediction of stock prices using sentiment analysis (Oh 2011). Researchers also examined the online corporate financial reports and found managers' opinions significantly correlate with companies' future performance (Li 2010).

If we take a more careful scrutiny of the sentiment analysis literature, it's not hard to discern that a large number of sentiment analysis problems were essentially formalized as a text classification problem (Liu 2012, Pang and Lee 2008). It's assumed that several sentimental tones such as positive, negative and neutral are embedded in the text from which such tones can be classified by a machine learning based classifier. Therefore, we believe the research problems shared by both sentiment analysis and text classification should be carefully addressed towards the way of making better sentiment analysis systems.

One important issue which is valid for both areas is the relative scarcity of labeled data and the ubiquity of unlabeled data. Although it's possible to gain more labeled data by devoting more resources, it's usually time-consuming and expensive to do so. Moreover, some particular labeling work needs to be done by domain experts to ensure the quality of the labels (e.g. labeling financial data). This makes the large scale labeling infeasible in practice. However, unlabeled textual data is not only ubiquitous on the Web but usually easy to obtain as well. This makes the semi-supervised learning paradigm, which is able to utilize both labeled and unlabeled data, attractive in many sentiment analysis and text classification tasks.

Since the time semi-supervised learning was proposed in text classification, there is a long term debate in the literature on the effectiveness of unlabeled data (McCallum and Nigam 1998, Nigam and McCallum, et al. 1998, Nigam and McCallum, et al. 2000, Cozman and Cohen, et al. 2003, Cohen and Cozman, et al. 2004, Zhang and Rudnicky 2006, Li and Zhou 2011). Though many rigorous and insightful results had been obtained, it seems the view on this problem has not been settled so far. We argue that the current view of this problem is largely limited by only considering the interplay of three dimensions, namely number of labeled data, number of unlabeled data and classification accuracy in the experiments. In order to have a more systematic understanding, an extra dimension has to be introduced. In this paper, we design a series of experiments to examine the interplay of these four dimensions. We show that the well-known ``performance degradation" issue caused by unlabeled data can be reproduced as a subset of the whole scenario. Meanwhile, our new view leads us to conclude that if the bias-variance trade-off is to be better balanced by a more effective feature selection mechanism, it's very likely that unlabeled data would help.

The structure of this paper is as follows. Section 2 summarizes the related work and points out the research gap. Section 3 presents the research methodology and the dataset. Section 4 illustrates the experiment results, presents and discusses the main findings. A feature selection framework which involves both labeled and unlabeled data is proposed in section 5. Discussions, limitations and future work are also presented in this section. Then the paper concludes with section 6.

\section{Related work and research gap}
Though it's widely accepted that the utilization of unlabeled data is desirable in many text classification tasks, the views on the effectiveness of unlabeled data on such tasks are quite diverse in the literature. McCallum and Nigam (1998) combined active learning with Expectation Maximization (EM) algorithm and showed this method is able to largely reduce the labeled data required in the training by utilizing more unlabeled data. Nigam and McCallum, et al. (1998) gave a theoretic analysis and showed that in the case of finite amount of label data and infinite amount of unlabeled data the positive value of unlabeled data in classification is definite. They then combined naive Bayes with EM algorithm to illustrate the usefulness of unlabeled data in text classification. In addition to many positive results, it was also showed in this study that unlabeled data may hurt in some situations. The authors attributed it to the violation of the model assumptions and addressed the issue by varying the weight of unlabeled data. Generally speaking, researchers possessed a relatively optimistic view on the utilization of unlabeled data in training text classifiers in these studies and some follow-up studies (Nigam and McCallum, et al. 2000, Toutanova and Chen, et al. 2001), although the performance degradation issue is acknowledged.

However, Cozman and Cohen, et al. (2003) indicated that the view on the effectiveness of unlabeled data may be too optimistic. They reviewed the literature and pointed out that the performance degradation issue caused by unlabeled data was actually not uncommon. Therefore, the usefulness of unlabeled data in text classification needs to be questioned and reexamined. They studied the asymptotic behavior of maximum likelihood estimation (MLE) and theoretically showed that when the model assumptions are correct, MLE is unbiased, thus unlabeled data is guaranteed to help in classification. However, when the model assumptions are incorrect, larger estimation bias is very likely to be introduced by unlabeled data. It's such bias causes the performance to degenerate. Since the model assumptions are almost always violated in practice (Nigam and McCallum, et al. 2000), such result is rather pessimistic to the utilization of unlabeled data.

While the performance degradation issue seems to be a hurdle in using unlabeled data, we observed a characteristic in most of the previous studies which may prevent us from understanding the performance degradation issue from a broader view. In these studies, only three variables, namely the interplay among the amount of labeled and unlabeled data and the classification accuracy were explicitly examined. Less effort was devoted to disclose how different vocabularies (features) influence the performance. We regard this as a research gap and will show in the experiments in this paper that this factor has a significant influence on the performance particularly in the presence of unlabeled data. Though other recent studies also tried to understand the effectiveness of unlabeled data in text classification (Zhang and Rudnicky 2006, Li and Zhou 2011), we observed that unlike our study, they either focused on the selection of unlabeled data or the improvement of the classification model itself. In addition, as we mentioned, many of the sentiment analysis problems were essentially text classification problems, thus the identified research gap is valid to sentiment analysis as well. We will further discuss the underlying motivations of carrying out the experiments in the financial sentiment analysis setting in the next section.

\section{Methodology and dataset}
\subsection{Research Methodology}
We chose to use multinomial naive Bayes to carry out the supervised learning part in the experiments. The reason why multinomial event model is a preferable choice in most of the text classification tasks was widely discussed in the literature (Ng and Jordan 2002, Nigam and McCallum 1998). 

Though there are a number of alternative methods to utilize unlabeled data in text classification, we observed that the combination of naive Bayes and EM algorithm was adopted in the previous seminal works in this field (Nigam and McCallum et al. 1998, Cozman and Cohen, et. al 2003). To keep our results relevant to these studies, we also adopted this method in our experiments. We briefly describe how EM algorithm is used in utilizing the unlabeled data in the following. 

Since the samples in unlabeled data don't have labels, therefore the log-likelihood function in the semi-supervised learning context is the following,

	\[
	\log l(\theta)=\log\prod\nolimits_{i=1}^{m_L} P(x^{(i)}, y^{(i)};\theta) + \log\prod\nolimits_{j=1}^{m_U} P(x^{(j)};\theta),
\]

\noindent in which $m_L$ is the number of the labeled samples, $m_U$ is the number of the unlabeled samples. Because the labels of the unlabeled data are invisible (or called hidden), we need to convert the right portion of the right-hand side of this equation to the following form by introducing a hidden variable $z^{(i)}$

\[
	\log l(\theta)=\log\prod\nolimits_{j=1}^{m_U} \sum\nolimits_{z^{(i)}} P(x^{(j)}, z^{(i)};\theta),
\]

\noindent in which $z^{(i)}$ is a random variable representing the hidden labels. It turns out that it's not tractable to use MLE to optimize the resulting equation because of the sum of log in it. That's why we need EM algorithm here.

There are two steps in the EM algorithm. The first step (the E-step) is to transfer the aforementioned log-likelihood function to a more tractable equation and find a tight lower bound for it. The second step (the M-step) is to optimize the tractable equation by MLE with respect to the parameters. Since EM is iterative by nature, we need to repeat the E-step and the M-step iteratively until the algorithm converges. It can be shown that EM is guaranteed to converge to a local optimum. More details of the algorithm can be found in the literature (Nigam and McCallum et al. 1998).

\subsection{Motivation of Using Financial Data}
The data we use in this paper are corporate financial reports which are publicly available online. There are both practical and theoretical motivations in choosing financial sentiment analysis as our experimental setting. In the practical side, the reason why applications of sentiment analysis are becoming prevalent these years is closely related to the rapid development of the Web and social networks. Since it's reasonable to anticipate more immense growth of the Web and social networks, we believe the impact of effective sentiment analysis systems will continue to grow in the future. Besides, sentiment analysis for finance is becoming one of the most prevailing settings. It distinguishes itself by having a number of influential applications including predicting future earnings (Li 2010) and detecting business frauds (Humpherys, et al. 2011), etc. 

There are also a few theoretical considerations in choosing corporate financial reports for our experiments. Firstly, multinomial naive Bayes imposes two theoretical assumptions to the data, namely 1) the data is generated by a naive Bayes mixture model; 2) the length of each classification unit doesn't have any influence to the real class labels. In our setting, every sentence within a certain section of the financial report is a classification unit and can be assigned a label, therefore our task is essentially a sentence level text classification task. As the variance of the length of sentences is significantly lower than the variance of the length of documents, the degree of violation of the second theoretical assumption is considerably alleviated. Since these two assumptions are almost always violated in document level classification in which performance degradation caused by unlabeled data was often observed, the immunity to one of the assumption violation in our setting gives us a chance to explore whether the simultaneous violation of the two assumptions is a necessary condition of the performance degradation issue. Secondly, it's a common practice in sentiment analysis literature to assume that each sentence expresses a single sentiment from a single opinion holder (Liu 2012). That means our task is not a multi-label sentiment analysis problem. Since most of the existing results were from the studies in the single-label text classification setting, we also would like to comply with such manner.

\subsection{Data Collection and Labeling}
We selectively collected textual data from the Management's Discussion and Analysis (MD\&A) section of Form 10-K documents. Form 10-K is an annual report required by U.S. Securities and Exchange Commission (SEC) in which the MD\&A section is dedicated for managers to discuss the operations of the company in detail. Managers' opinions and judgments of the company are usually presented in this section.

In order to make our test results have a good chance to approximate the generalization error, the collected data needs to be comparable to the random samples. We adopted the following procedure in the data collection. We randomly selected 10 industries and then selected only one company within each industry. The reason to do so is to avoid the possible similar writing habits within the same industry. By having the company list, we downloaded their financial reports for the year of 2011 and 2012 from the SEC website and extracted the MD\&A section in the preprocessing. We assumed every sentence can be categorized into one of the three sentimental tones, namely positive, negative and neutral. This is consistent with the common practice in many previous sentiment analysis studies (Pang and Lee 2008). 

\begin{table}[h]
	\centering
		\scalebox{.94} {\begin{tabular}{ c | c | c }
			\hline
			\textbf{Company} & \textbf{Industry} & \textbf{Labeled Data Stat} \\ \hline \hline			
			\begin{small}Accenture (Ac)\end{small}  & \begin{small}Consulting\end{small} & \begin{small}356 (39\%\/21\%\/40\%)\end{small} \\ \hline
			\begin{small}AECOM (Ae)\end{small} & \begin{small}Engineering\end{small} & \begin{small}396 (20\%\/17\%\/63\%)\end{small} \\  \hline
			\begin{small}AXA (Ax)\end{small} & \begin{small}Insurance\end{small} & \begin{small}695 (31\%\/18\%\/51\%)\end{small} \\  \hline
			\begin{small}Cisco (Cs)\end{small} & \begin{small}Telecom\end{small} & \begin{small}583 (37\%\/26\%\/37\%)\end{small} \\  \hline
			\begin{small}Coach (Co)\end{small} & \begin{small}Luxury Goods\end{small} & \begin{small}328 (33\%\/8\%\/59\%)\end{small} \\  \hline
			\begin{small}Dell (De)\end{small} & \begin{small}Manufacturing\end{small} & \begin{small}587 (27\%\/8\%\/65\%)\end{small} \\  \hline
			\begin{small}Ford (Fd)\end{small} & \begin{small}Automobile\end{small} & \begin{small}646 (34\%\/15\%\/51\%)\end{small} \\  \hline
			\begin{small}Microsoft (Ms)\end{small} & \begin{small}Software\end{small} & \begin{small}383 (22\%\/19\%\/59\%)\end{small} \\  \hline
			\begin{small}Monsanto (Mon)\end{small} & \begin{small}Agriculture\end{small} & \begin{small}459 (35\%\/10\%\/55\%)\end{small} \\  \hline
			\begin{small}Morgan Stanley (Mor)\end{small} & \begin{small}Banking\end{small} & \begin{small}812 (17\%\/14\%\/69\%)\end{small} \\  \hline
		\end{tabular}}
	\caption{Selected companies and labeled data statistics}
	\label{tab:table1}
\end{table}

We then labeled all the sentences of the year 2012 and put them into our pool of labeled data. The data from the year 2011 was put into the pool of unlabeled data. In order to ensure the quality of the labels, the labeling work was directed and double checked by a Certified Public Accountant (CPA). The final labeled data pool contains 5245 sentences and the unlabeled data pool contains 5342 sentences. Table 1 summaries the industries and companies we chose. The statistics of the labeled data is also presented in the table. The right most column not only shows the amount of labeled sentences from each company, it also shows the percentage of the positive, negative and neutral sentences within the data respectively. We can see from the table that the data from each company seems to possess different characteristics. We regard it as a positive sign in approximating random samples.

\section{Experiments}
\begin{figure*}
  \centering
  \subfigure[Labeled data from Ms]{
    \label{fig:subfig:a} 
    \includegraphics[scale=0.29]{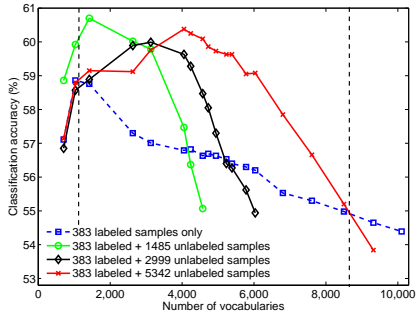}}
  \subfigure[Labeled data from Ms, Co]{
    \label{fig:subfig:b} 
    \includegraphics[scale=0.29]{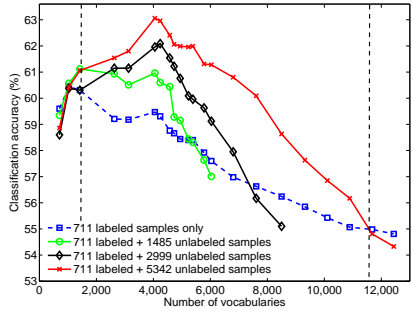}}
		\subfigure[Labeled data from Ms, Co, De]{
    \label{fig:subfig:c} 
    \includegraphics[scale=0.29]{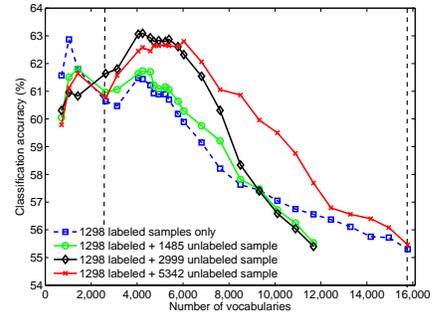}}
		\subfigure[Labeled data from Ms, Co, De, Mon]{
    \label{fig:subfig:d} 
    \includegraphics[scale=0.29]{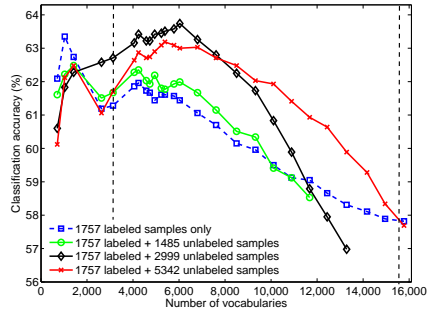}}
		\subfigure[Labeled data from Ms, Co, De, Mon, Ae]{
    \label{fig:subfig:e} 
    \includegraphics[scale=0.29]{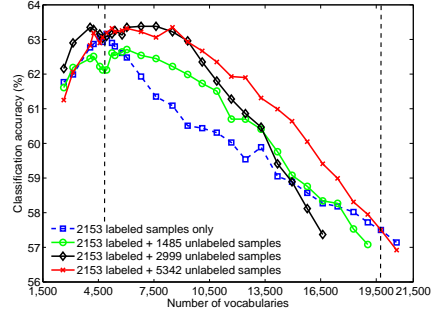}}
  \caption{Experiment one.}
  \label{fig:fig1} 
\end{figure*}

\begin{figure*}
  \centering
  \subfigure[Labeled data from Ms]{
    \label{fig:subfig:a} 
    \includegraphics[scale=0.29]{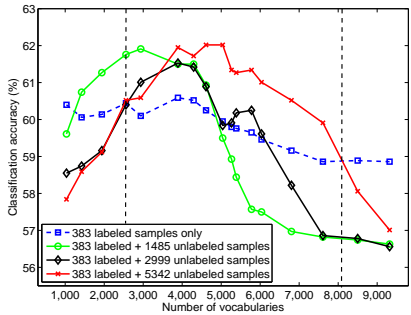}}
  \subfigure[Labeled data from Ms, Mon]{
    \label{fig:subfig:b} 
    \includegraphics[scale=0.29]{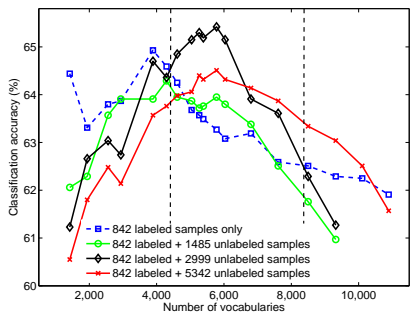}}
		\subfigure[Labeled data from Ms, Mon, Mor]{
    \label{fig:subfig:c} 
    \includegraphics[scale=0.29]{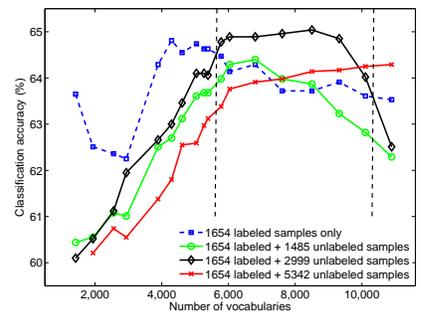}}
		\subfigure[Labeled data from Ms, Mon, Mor, Cs]{
    \label{fig:subfig:d} 
    \includegraphics[scale=0.29]{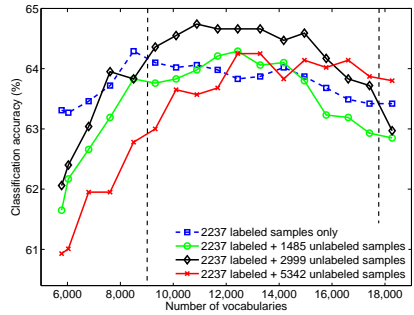}}
		\subfigure[Labeled data from Ms, Mon, Mor, Cs, Ac]{
    \label{fig:subfig:e} 
    \includegraphics[scale=0.29]{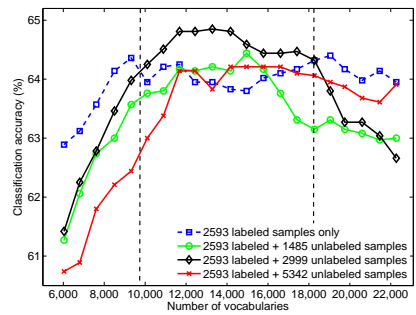}}
  \caption{Experiment two.}
  \label{fig:fig2} 
\end{figure*}

\subsection{Why Design The Experiment This Way}
We previously mentioned that most of the existing studies tend to understand the usefulness of unlabeled data by viewing the interplay of three factors namely the amount of labeled data, the amount of the unlabeled data and the classification performance. It was concluded that in addition to the classifier itself, both the amount of labeled and unlabeled data would change the dynamics of the bias-variance trade-off and therefore influence the classification performance. The performance degradation issue was also observed with the increase of the amount of labeled data in the presence of the unlabeled data (Cozman and Cohen, et. al 2003). However, we argue that there is one additional factor, the choice of features, has the potential to significantly influence the bias-variance trade-off. For instance, if the feature selection procedure is not to be carefully conducted every time we vary the amount of labeled and unlabeled data, the balance between bias and variance is likely to be broken. Due to this reason, the ensuing classification may suffer from this imbalance. Since the introduction of the feature selection procedure was either not explicitly mentioned or largely simplified in some of the previous seminal works in this field (Nigam and McCallum et al. 1998, Nigam and McCallum et al. 2000, Cozman and Cohen, et. al 2003, Cohen and Cozman, et al. 2004), it's not entirely clear whether the researchers performed feature selection every time they varied the amount of labeled or unlabeled data. Thus, there are reasons to devise a mechanism to further examine the consequences of \textit{not} doing this.

\subsection{Experimental Procedure}
In this study, we considered the choice of features as one of the dimensions in the experiments. We used the following procedure to vary the amount of labeled data, the amount of unlabeled data and the choice of features simultaneously to see how they would collectively influence the classification performance. In order to better interpret the results and fit the results into a 2D figure, we fixed the amount of labeled data in each subfigure of figure 1 and figure 2 within which we varied the amount of unlabeled data and the choice of features. We then carried out multiple trials with gradually increasing amount of labeled data. Therefore, we were able to generate a series of subfigures which collectively visualized the interplay of all the considered factors. In order to gain confidence on the generalizability of our results, we randomly divided our labeled data into the training set and the test set twice and performed the aforementioned experimental procedure for each choice of data division respectively. The detailed choice of the labeled training data and the sequences of adding the labeled data into the training set are described in the captions of figure 1 and figure 2.

We varied the amount of unlabeled training data in exactly the same way for each fixed amount of labeled training data. The procedure is the following. We added unlabeled data in the training three times. The first time included 1485 unlabeled samples from Co, De and Ae. The second time include 2999 unlabeled samples from Co, De, Ae, Fd, Ax and Ms. The last time included 5342 unlabeled samples from all the companies. This sequence was decided by a random choice. The only purpose of this procedure was to vary the amount of the unlabeled data in an unbiased way. The test set for all the subfigures within each figure are the same in which contained all the available samples.

We varied the number of vocabularies in the following. Firstly, we created a list of words by scanning all the data we have. Numbers, punctuation and other non-characters were removed from the list. When increasing the number of vocabularies, the ones came from the labeled training and test set was firstly added to the dictionary at a random sequence. When we need more vocabularies, the ones came from the unlabeled training set was then randomly added to the dictionary. In the case the number of vocabularies is still not sufficient, a standard English word list was used to add more novel vocabularies.

\subsection{Analysis of The Findings}
There are a few interesting findings in the experiments. First of all, all the curves, including both the ones for labeled training data only and the ones use both labeled and unlabeled training data, appear a performance increase and then follow by a performance decrease. Since we mentioned several reasons to believe the results from our test sets approximate the generalization error to a certain extent, we conclude the best performance of each curve is a product of the well balanced bias-variance trade-off which is achieved by choosing a satisfactory vocabulary number in the dictionary. 

Secondly, we can always find an interval of appropriate amount of vocabularies in the diagrams in which the classification performance is always increased by adding unlabeled data in training the classifier. We marked the longest intervals of this kind in each subfigure by vertical dotted lines. 

Thirdly, the best achieved performance is always the result of using both labeled and unlabeled data in training. In the cases where the performance achieved by the usage of unlabeled data is not as good (e.g. the red curve in figure 2 (b)), the performance is still comparable.

The first three findings imply that if the bias-variance trade-off is well balanced, by choosing the number of vocabularies well, the usage of unlabeled data often leads to superior classification performance. Also, we didn't see any significant performance degradation in the cases where the performance is not superior, given the number of vocabularies is appropriate.

The fourth finding is when the bias-variance trade-off is not well addressed in the presence of unlabeled data, the drop in performance is rather radical. It's almost certain from the experiments that the velocity of such a drop is significantly higher than that in the case with only labeled training data. By considering all the findings so far, it leads us to the following reasoning. It seems there is always an interval of vocabulary number in which significant performance degradation can be observed with the increasing usage of unlabeled data. Meanwhile, given all the other settings the same, significant performance improvement can also be observed by just varying the vocabulary numbers to a more appropriate interval. What is worth mentioning is this phenomenon won't go away with the increase of labeled training data, at least according to our experiments. Figure 3 illustrates this point visually.

\begin{figure}
  \centering
  \subfigure[]{
    \label{fig:subfig:a} 
    \includegraphics[scale=0.21]{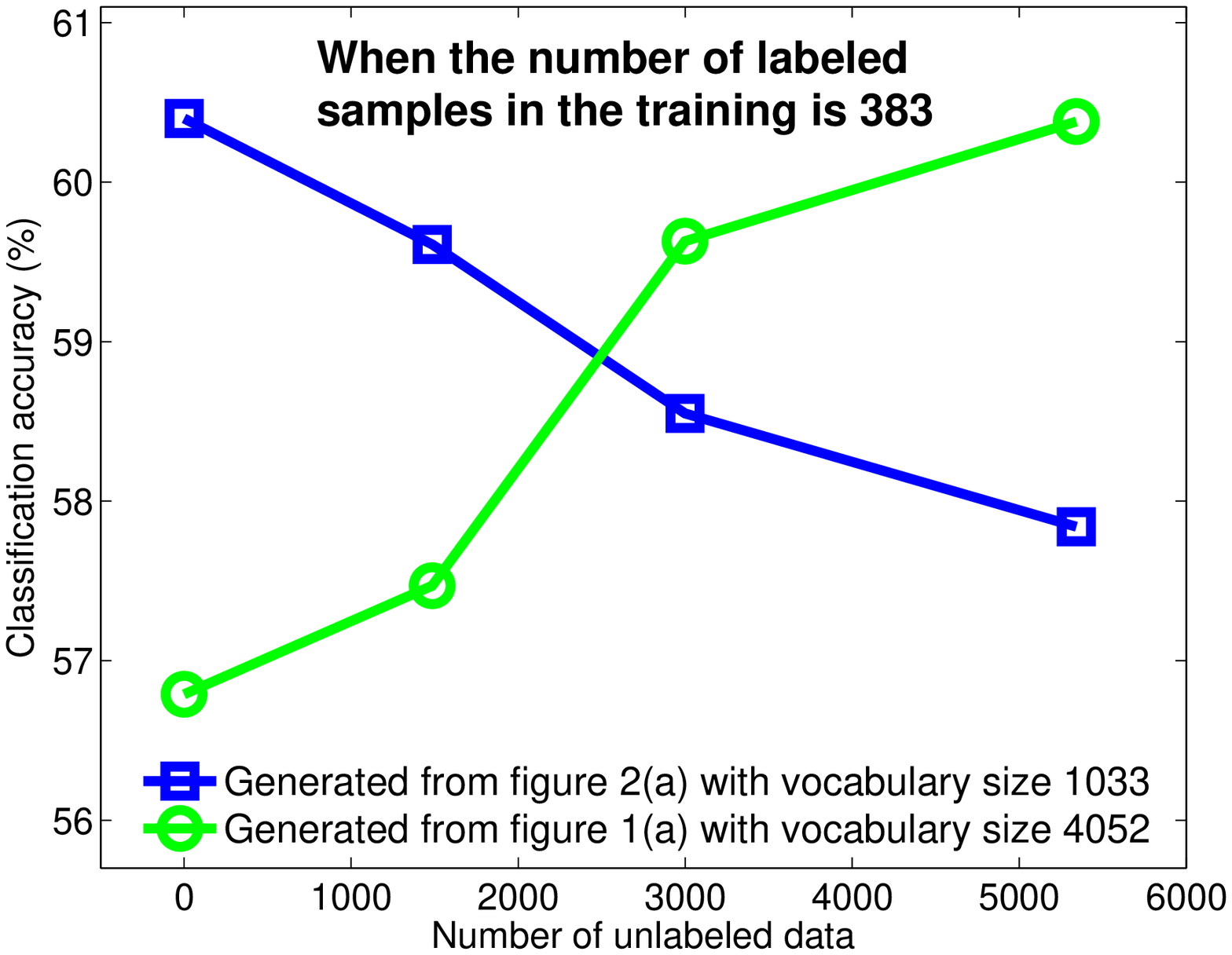}}
  \subfigure[]{
    \label{fig:subfig:b} 
    \includegraphics[scale=0.21]{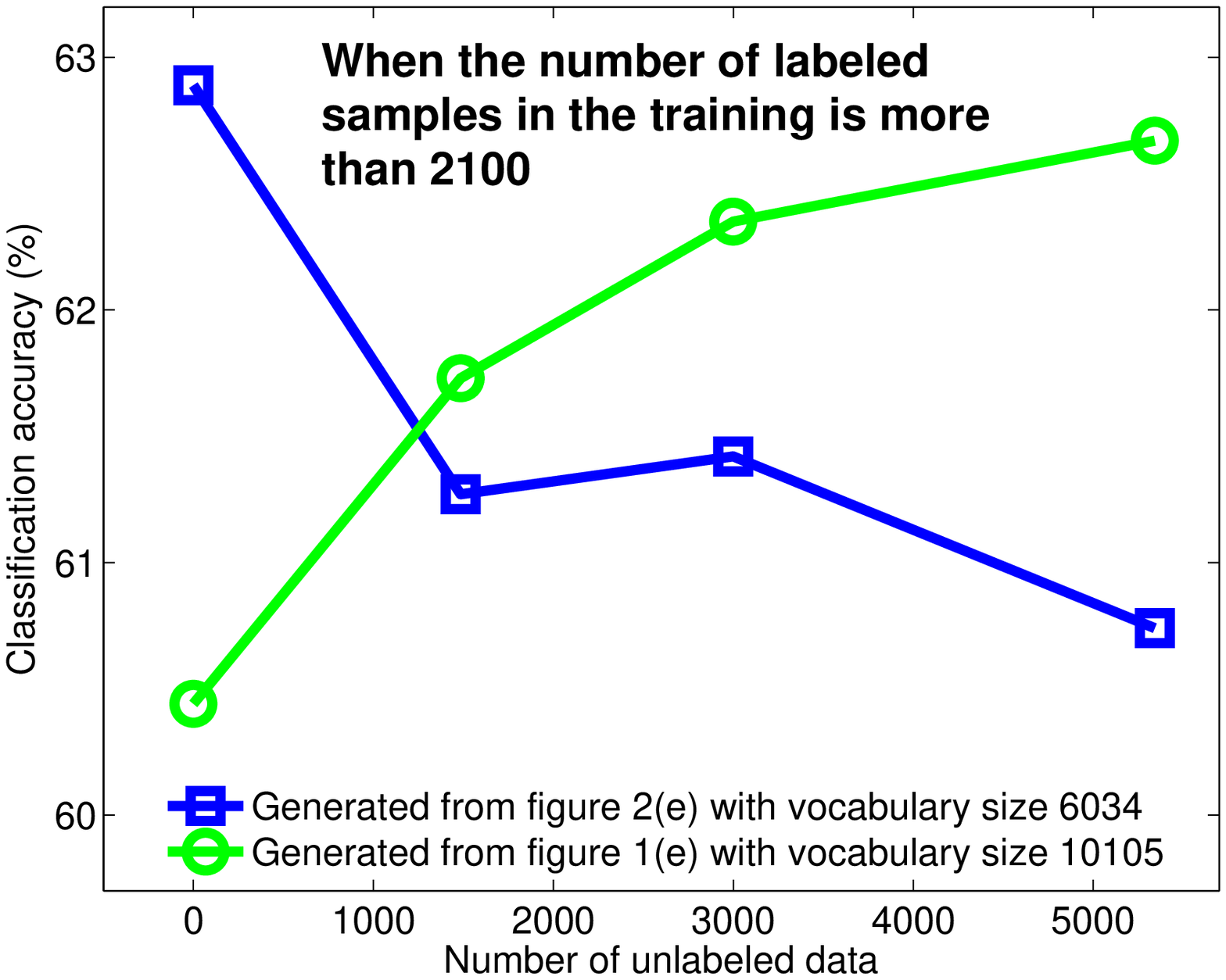}}
  \caption{No matter the amount of labeled samples is big or small, performance degradation and performance improvement can both be observed by manipulating the bias-variance trade-off.}
  \label{fig:fig3} 
\end{figure}

Figure 3(a) shows when the amount of labeled data is small in training, both performance degradation and performance improvement can be observed if we increase of the usage of unlabeled data. Figure 3(b) shows the same pattern can also be observed even if we use significantly more labeled training data compare to the case in figure 3(a). Though we can not be sure without a prove, it seems our experiment results suggest that the well-known performance degradation issue caused by unlabeled data may be a subset of the whole scenarios we showed. This implies if we perform an effective feature selection procedure during the training, we can largely avoid the performance degradation when using unlabeled data in text classification. However, we claim this feature selection procedure won't success if we only consider the labeled data. We discuss why this is the case in the next section and propose an alternative feature selection framework.

\section{Discussion}
\subsection{An Alternative Feature Selection Framework}
We now discuss why we should consider both labeled and unlabeled data in the feature selection procedure. We can see from both figure 1 and figure 2 that the best classification performance is usually achieved by different choices of vocabulary numbers. For instance, in figure 2(a), there are considerable offsets among these optimal points. However, if we only use labeled data in the feature selection, only one ``appropriate" vocabulary number can be chosen regardless how much unlabeled data we used. This implies that the bias-variance trade-off is likely to be broken when we introduce unlabeled data to the training if we do feature selection in the conventional way. On the other hand, this further implies that even rigorous feature selection procedure was carried out in the previous studies, performance degradation issue may also be prevalent if only labeled data was considered in the process. However, this is a weak claim. More experiments need to be done to support this.

In the light of the previous discussion, we propose that unlabeled data needs to be included in the feature selection procedure. Since it's a common assumption that labels are not absent but just invisible for the unlabeled data, what we need is to find a mechanism to reveal the labels. One possible way to use unlabeled data in the feature selection is to use the pseudo labels generated by the EM algorithm used in this paper. However, this may not be the most effective method for feature selection. Since more detailed discussion on this problem is not the focus of this paper, we will leave it to the future study.

\subsection{More discussions, Limitations and Future Work}
The findings in this paper contribute to the text classification literature by providing a way to look at the puzzling performance degradation issue in a different and possibly broader view. We believe this is an important issue to address because the ability of utilizing unlabeled data in text classification will only be more attractive in the future since the rapid growth of the Web is inevitable. Such results should also be of interest to sentiment analysis community because we anticipate more text classification based sentiment analysis applications to appear in the near future.

However, there are a few limitations in this study. First of all, though we tend to avoid any bias in the data collection, data processing and the experiments, we only carried out the experiments in one dataset. It's possible that the findings generated by our experiments were biased by some special characteristics of financial data. Secondly, our setting is a sentence level text classification problem. The generalizability of the findings in the document level text classification problems needs to be further examined.

There are also a few interesting future research directions. It would be of very high interest that a prove on whether our results could generalize can be given. Detailed study on feature selection in the presence of unlabeled data is another promising topic. We also mentioned previously that unlabeled data seems to be more sensitive to the bias-variance trade-off. The disclosure of the underlying reasons would be very interesting. In addition, it seems the immunity to the violation of the text length assumption doesn't prevent the performance degradation from happening. The way how the data generating assumption influences the performance in the presence of unlabeled data is worth exploring as well. Finally, it seems it's not guaranteed that more unlabeled data would always lead to better performance even if the bias-variance trade-off is well addressed. Therefore, the selection of the unlabeled data is an interesting topic.

\section{Concluding remarks}
In this paper, we examined the utilization of unlabeled data in text classification by using well prepared financial textual data. We added an additional dimension in the experiments and showed that the performance degradation issue caused by unlabeled data may be a subset of a broader scenario. We carefully discussed such scenario and concluded from the experiment results that a feature selection procedure which considers both labeled and unlabeled data is a promising candidate in avoiding the performance degradation in semi-supervised text classification. The implications of this study for both text classification and sentiment analysis were also discussed.

\section{ Acknowledgments}
Research reported in this paper was partially supported by Innovation and Technology Fund of Hong Kong SAR government (Project ITP/021/09AP and ITP/036/10AP). We thank the anonymous reviewers for their valuable comments and suggestions. 

\section{Reference}
\begin{small}
Cohen, I., F. G. Cozman, et al. 2004. Semisupervised Learning of Classifiers: Theory, Algorithms, and Their Application to Human Computer Interaction. \textit{IEEE Transactions on Pattern Analysis and Machine Intelligence} 26(12): 1553--1567.
\smallskip \noindent \\
Cozman, F. G., I. Cohen, et al. 2003. Semi-Supervised Learning of Mixture Models. \textit{In Proceedings of ICML}, 99--106. 
\smallskip \noindent \\
Humpherys, S. L., K. C. Moffitt, et al. 2011. Identification of fraudulent financial statements using linguistic credibility analysis. \textit{Decision Support Systems} 50(3): 585--594. 
\smallskip \noindent \\
Li, F. 2010. The Information Content of Forward-Looking Statements in Corporate Filings��A Naive Bayesian Machine Learning Approach. \textit{Journal of Accounting Research} 48(5): 1049--1102. 
\smallskip \noindent \\
Li, Y.-F. and Z.-H. Zhou 2011. Towards Making Unlabeled Data Never Hurt. \textit{In Proceedings of ICML}, 1081--1088. 
\smallskip \noindent \\
Liu, B. 2012. Sentiment Analysis and Opinion Mining, \textit{Morgan \& Claypool Publishers}.
\smallskip \noindent \\
McCallum, A. K. and K. Nigam 1998. Employing EM and Pool-Based Active Learning for Text Classification. \textit{In Proceedings of ICML}, 359--367.
\smallskip \noindent \\
Andrew Y. Ng, Michael I. Jordan 2002. On discriminative vs. generative classifiers: A comparison of logistic regression and naive bayes. \textit{In Proceedings of NIPS}, 841--848
\smallskip \noindent \\
Nigam, K., A. McCallum, et al. 1998. Learning to Classify Text from Labeled and Unlabeled Documents. \textit{In Proceedings of AAAI}, 792--799.
\smallskip \noindent \\
Nigam, K., A. K. McCallum, et al. 2000. Text Classification from Labeled and Unlabeled Documents using EM. \textit{Machine Learning} 39(2-3): 103--134.
\smallskip \noindent \\
Oh, C. and O. Sheng 2011. Investigating Predictive Power of Stock Micro Blog Sentiment in. Forecasting Future Stock Price Directional Movement. \textit{In Proceedings of International Conference on Information Systems (ICIS 2011)}.
\smallskip \noindent \\
Pang, B. and L. Lee 2008. Opinion Mining and Sentiment Analysis. \textit{Foundations and Trends in Information Retrieval} 2(1-2): 1--135.
\smallskip \noindent \\
Toutanova, K., F. Chen, et al. 2001. Text Classification in a Hierarchical Mixture Model for Small Training Sets. \textit{In Proceedings of CIKM}, 105--113.
\smallskip \noindent \\
Zhang, R. and A. I. Rudnicky 2006. A New Data Selection Principle for Semi-Supervised Incremental Learning. \textit{In Proceedings of ICPR}, 780-783.
\end{small}

\end{document}